\documentclass[letterpaper]{article} 
\usepackage{aaai25}  
\usepackage{times}  
\usepackage{helvet}  
\usepackage{courier}  
\usepackage[hyphens]{url}  
\usepackage{graphicx} 
\usepackage{natbib}  
\usepackage{caption} 
\usepackage{amsmath}
\usepackage{amsfonts}
\usepackage{algorithm}
\usepackage{algorithmic}
\usepackage{multirow}
\usepackage{booktabs}
\usepackage{soul}
\usepackage{bbding}

\usepackage{newfloat}
\usepackage{listings}
\DeclareCaptionStyle{ruled}{labelfont=normalfont,labelsep=colon,strut=off} 
\floatstyle{ruled}
\newfloat{listing}{tb}{lst}{}
\floatname{listing}{Listing}
\title{MUC: Mixture of Uncalibrated Cameras for\\ Robust 3D Human Body Reconstruction}
\author{
    Yitao Zhu\textsuperscript{\rm 1}\equalcontrib,
    Sheng Wang\textsuperscript{\rm 2, \rm 3}\equalcontrib,
    Mengjie Xu\textsuperscript{\rm 1},
    Zixu Zhuang\textsuperscript{\rm 2, \rm 3},\\
    Zhixin Wang\textsuperscript{\rm 1},
    Kaidong Wang\textsuperscript{\rm 1},
    Han Zhang\textsuperscript{\rm 1, \rm 4},
    Qian Wang\textsuperscript{\rm 1, \rm 4}
}
\affiliations{
    \textsuperscript{\rm 1}School of Biomedical Engineering \& State Key Laboratory of \\Advanced Medical Materials and Devices, ShanghaiTech University\\
    \textsuperscript{\rm 2}School of Biomedical Engineering, Shanghai Jiao Tong University\\
    \textsuperscript{\rm 3}Shanghai United Imaging Intelligence Co., Ltd.\\
    \textsuperscript{\rm 3}Shanghai Clinical Research and Trial Center\\
}

\usepackage{bibentry}

\begin{document}

\maketitle

\begin{abstract}
Multiple cameras can provide comprehensive multi-view video coverage of a person. Fusing this multi-view data is crucial for tasks like behavioral analysis, although it traditionally requires camera calibration—a process that is often complex. Moreover, previous studies have overlooked the challenges posed by self-occlusion under multiple views and the continuity of human body shape estimation.
In this study, we introduce a method to reconstruct the 3D human body from multiple uncalibrated camera views. Initially, we utilize a pre-trained human body encoder to process each camera view individually, enabling the reconstruction of human body models and parameters for each view along with predicted camera positions.
Rather than merely averaging the models across views, we develop a neural network trained to assign weights to individual views for all human body joints, based on the estimated distribution of joint distances from each camera.
Additionally, we focus on the mesh surface of the human body for dynamic fusion, allowing for the seamless integration of facial expressions and body shape into a unified human body model.
Our method has shown excellent performance in reconstructing the human body on two public datasets, advancing beyond previous work from the SMPL model to the SMPL-X model. This extension incorporates more complex hand poses and facial expressions, enhancing the detail and accuracy of the reconstructions. Crucially, it supports the flexible ad-hoc deployment of any number of cameras, offering significant potential for various applications. Our code is available at \url{https://github.com/AbsterZhu/MUC}.
\end{abstract}

%

\section{Introduction}
\label{sec:intro}

Recently, there has been a growing interest in creating 3D whole-body models from 2D images, especially for virtual reality and 3D animation. 
This process, known as \textit{expressive whole-body mesh recovery}, combines the estimation of 3D human body pose, hand gesture, and facial expression. The progress is remarkable in single camera (monocular) setting such as SMPLer-X \cite{cai2023smpler}, which has become a foundation model in this field.
Yet using a single camera still faces challenges such as occlusions, which can severely degrade the reconstruction quality. 



Utilizing multiple camera views, multi-view videos provide essential visual cues from various angles and are widely used in applications like behavioral studies, where they facilitate the detailed observation of children's behaviors in natural settings \cite{ballan2010unstructured, schmidt2008effects}. While the first step in many traditional approaches involves spatial camera calibration \cite{he2020epipolar}, this process is often cumbersome, requiring specialized tools and complex optimizations. Such time-consuming requirements hinder the efficient and convenient deployment of multi-camera setups across numerous sites. In response, this paper proposes a novel method for fusing data from multiple uncalibrated cameras to reconstruct 3D human bodies more effectively.


Many multi-view methods build on single-view approaches, typically by averaging predictions from individual views to estimate 3D human body pose and shape, which can lead to incoherence and reduced accuracy \cite{kanazawa2018end,kolotouros2019learning,jafarian2021learning}. Others map features from multiple views onto a common space for fusion \cite{yu2022multiview}, requiring complex coordinate transformations that increase method complexity. In contrast, the pixel-aligned feedback fusion method \cite{jia2023delving} iteratively refines estimations through aligned mesh vertex features, boosting accuracy without complex spatial mappings but necessitating multiple iterations. Additionally, the scarcity of multi-view data compared to single-view data complicates the direct training of multi-view models for optimal performance.


In this study, we introduce the Mixture of Uncalibrated Cameras (MUC), a novel pipeline for 3D human body reconstruction from multiple uncalibrated camera views, combining single-view encoding and multi-view fusion. Recognizing that early single-view methods effectively capture the relationship between observation angle and human pose, we apply a proven single-view model to each camera to independently reconstruct body parts. Notably, parts captured closer and without occlusion are more accurately reconstructed due to fewer self-occlusions. Therefore, we assess the quality of these single-view reconstructions, dynamically weighting key body parts across views to optimally fuse multi-view data.

It is critical for our method to assess the quality of the single-view reconstuctions. 
To this end, we propose two strategies.
\textbf{(1) Joint Reweighting.} For all body joints predicted from each view, we use a joint reweighting network to acquire the weights of the human body joints based on the predicted camera positions. 
We also implement a distance-based distribution loss function, aiming to reduce the weights of body parts that have high uncertainty in reconstruction. 
\textbf{(2) Surface Reweighting.} To integrate facial expression with body smoothly, we design a surface reweighting network. 
This network utilizes UV map, corresponding to vertex normals of the mesh of human body, to estimate dynamic weighting toward seamless combination of facial expressions and body shape.

\textbf{Contributions}: The contributions of this work can be summarized in three items.
\begin{itemize}
\item[(1)] We present a groundbreaking multi-view method for reconstructing human pose, shape and facial expression.
Our method does not require calibration or geometric mapping among views.
It is scalable to an arbitrary number of cameras (including single view) out of the box.
\item[(2)] We develop a novel distribution prediction strategy based on distance to assess the joint importance from various viewpoints, coupled with surface reweighting for continuous fusion of facial expression and body shape.
\item[(3)] Our method establishes a new benchmark in calibration-free 3D human body reconstruction, surpassing existing state-of-the-art approaches and expanding upon previous work by transitioning from the SMPL human model to the more detailed SMPL-X model.
\end{itemize}

\section{Related Work}
\label{sec:relate}
Rising demand for 3D human body reconstruction has catalyzed a division in research into single-view and multi-view methods. Single-view approaches, bolstered by richly annotated datasets, have reached high levels of accuracy but often falter when dealing with occlusions. Conversely, multi-view methods excel at managing occlusions but are hindered by the scarcity of multi-view datasets (i.e., datasets with multiple observation angles), leading to ongoing research aimed at improving feature fusion techniques in these frameworks.

\textbf{Single-View Human Body Reconstruction.} 
In the context of 3D human body reconstruction from a single viewpoint, models like SMPL~\cite{loper2023smpl} and SMPL-X~\cite{pavlakos2019expressive} provide effective low-dimensional parameterizations for human poses and shapes. Despite numerous applications using these models \cite{feng2021collaborative,choutas2020monocular,rong2021frankmocap,zhou2021monocular}, resolution constraints and limited field-of-view often hinder accurate facial and finger estimation. To address this, some methods, such as ExPose~\cite{choutas2020monocular}, incorporate body-driven attention mechanisms and refinements from specialized datasets for enhanced detail, while OSX~\cite{lin2023one} presents a unified end-to-end model for comprehensive pose and shape estimation, achieving remarkable outcomes. Extensive datasets, both real and synthesized \cite{lin2014microsoft,patel2021agora,huang2022capturing,ionescu2013human3}, and targeted non-full-body datasets \cite{lin2023one} are crucial in overcoming occlusion challenges in single-view reconstruction. SMPLer-X~\cite{cai2023smpler} leverages 4.5 million images from 32 datasets to become the state-of-the-art foundation model in this domain.

\textbf{Multi-View Human Body Reconstruction}

The primary approach to multi-view reconstruction employs calibration parameters (intrinsic and extrinsic), as demonstrated by SMPLify-X~\cite{pavlakos2019expressive}, which reduces 2D keypoint and silhouette reprojection errors in a unified system~\cite{li20213d}. Although effective, this method's heavy reliance on precise camera calibration can be impractical in dynamic environments.

When camera parameters are not available, alternative methods are used. A simple approach is averaging results from single-view models to achieve multi-view fusion. More complex learning-based methods include warping feature maps to align views without calibration~\cite{qiu2019cross} and fusing images by mapping features onto a semantic model of the human body, utilizing self-attention to integrate features~\cite{yu2022multiview}. These methods manage to bypass traditional calibration but can struggle with complex mappings and self-occlusion.
The pixel-aligned feedback fusion technique iteratively refines body parameters by aligning features on mesh vertices. While this enhances accuracy, the multiple iterations required reduce computational efficiency, potentially limiting its suitability for real-time applications~\cite{jia2023delving}.
Overall, these methods depend on the SMPL model, which falls short in detailing hand movements and facial expressions, underscoring the need for advanced models like SMPL-X to better tackle these issues.


\begin{figure*}[h]
  \centering
  \includegraphics[width=0.95\textwidth]{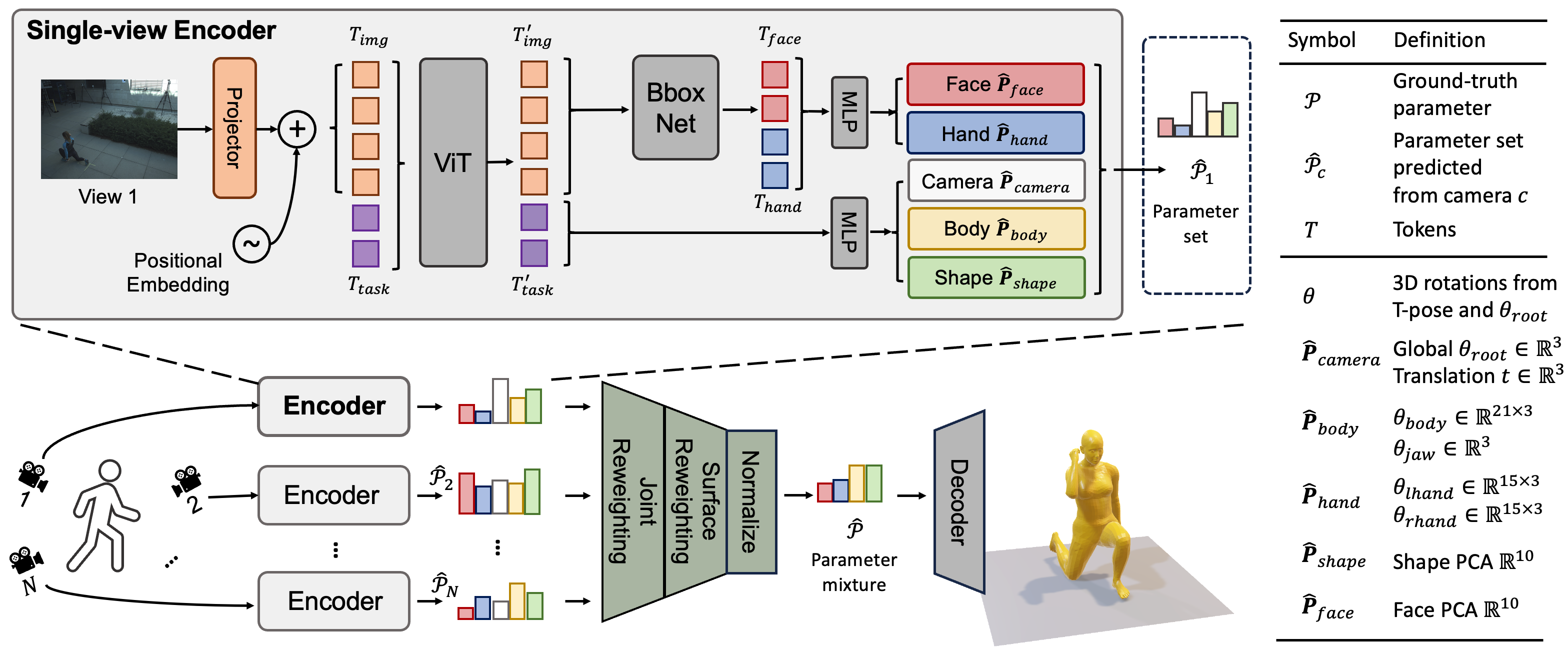}
  \caption{Diagram of Mixture of Uncalibrated Cameras (MUC), with the architecture of the single-view encoder shown on the top. On the right, a notation table lists the symbols and their definitions used in the model of human body.}
  \label{fig:method}
\end{figure*}

\section{Method}
\label{sec:method}
Our work introduces an innovative approach to reconstruct 3D human body using multi-view images from multiple uncalibrated cameras as shown in Figure \ref{fig:method}. 
We adopt a pre-trained single-view Vision Transformer (ViT) based foundation model SMPLer-X \cite{cai2023smpler} as encoder for each view.
The core of our approach comprises two key reweighting networks: a Joint Reweighting Network (JRN) and a Surface Reweighting Network (SRN). 
They cleverly figures out which parts of the images are reliable and use this information to reconstruct better joints and surface. 

The sequence of reweighting first the joints and then the surface is intentional, as the surface mesh is contingent upon the underlying joint-based skeleton structure. 
Upon the completion of the reweighting processes, we normalize and amalgamate all reweighted parameters into a unified parameter set. The parameter set is then fed into a decoder, which reconstructs the 3D human body.

\subsection{Single-view Encoder}
To exploit the benefits of extensive datasets in single-view 3D human body reconstruction, we adopt a pre-trained encoder from SMPLer-X, a generalist foundation model specialized for monocular tasks. The encoder is portrayed in the top of Figure \ref{fig:method}. 

We introduce how the single-view encoder works briefly, which is important to subsequent multi-view fusion. 
First, a human body image is split into a sequence of fixed-size non-overlapping patches, which are linearly projected to image tokens $T_{img}$. They are then concatenated with learnable task tokens \( T_{task} \), which are transformed by ViT to \(T^{\prime}_{img}\) and \(T^{\prime}_{task}\). A BboxNet module predicts bounding boxes to localize face and hands in the feature map. Subsequently, MLPs are employed to regress a set of parameters \(\hat{\mathcal{P}} = \{\hat{\mathbf{P}}_{body}, \hat{\mathbf{P}}_{hand}, \hat{\mathbf{P}}_{shape}, \hat{\mathbf{P}}_{face}, \hat{\mathbf{P}}_{camera}\}\). 
Notably, $\hat{\mathbf{P}}_{body}, \hat{\mathbf{P}}_{hand}$ are joints parameter in rotation $\theta$, $\hat{\mathbf{P}}_{shape}, \hat{\mathbf{P}}_{face}$ are surface parameter in PCA top-10 components to control the curvature of different parts of the mesh.
This encoder is tailored to minimize the difference between the estimated parameters \(\hat{\mathcal{P}}\) and the ground-truth \(\mathcal{P}\), focusing particularly on areas rich in details such as the hands and face.
Further, a SMPL-X layer, as described in \cite{pavlakos2019expressive}, can be the decoder to derive the 3D mesh of human body from \(\hat{\mathcal{P}}\).

\subsection{Multi-view Joint Reweighting}
\label{sec:jir}

In the last section, we brief the single-view encoder. In this section, we introduce how the proposed Joint Reweighting Network (JRN) can mix the joint parameters from multiple uncalibrated cameras.

\textbf{Network architecture} The JRN employs two MLPs to mix joint landmark \(\hat{\mathbf{P}}_{body, c}\) and hand landmark \(\hat{\mathbf{P}}_{hand, c}\) based on camera position \(\hat{\mathbf{P}}_{camera, c}\) for the \(c\)-th camera.
\begin{itemize}
    \item The first MLP$_{body}$ gets task token $T'_{task}$ and camera parameter $\hat{\mathbf{P}}_{camera, c}$, and outputs the score $s_c^{body} = \text{MLP}_{body}(T^{\prime}_{task, c}, \hat{\mathbf{P}}_{camera, c})$.
    And $\hat{\mathbf{P}}_{body}$ is mixed as
    \begin{equation}
    \label{equ:P_Body}
        \hat{\mathbf{P}}_{body} = \frac{\sum_{c=1}^{N}\hat{\mathbf{P}}_{body, c}\cdot s_c^{body}}{\sum_{c=1}^{N}s_c^{body}}.
    \end{equation}

    \item The second MLP$_{hand}$ gets hand token $T_{hand, c}$ and camera parameter $\hat{\mathbf{P}}_{camera, c}$. The hand token is localized and cropped from image token by BboxNet as shown in Figure~\ref{fig:method}: $s_c^{hand} = \text{MLP}_{hand}(T_{hand, c}, \hat{\mathbf{P}}_{camera, c})$.
    And $\hat{\mathbf{P}}_{hand}$ is mixed as
    \begin{equation}
    \label{equ:P_Body1}
        \hat{\mathbf{P}}_{hand} = \frac{\sum_{c=1}^{N}\hat{\mathbf{P}}_{hand, c}\cdot s_c^{hand}}{\sum_{c=1}^{N}s_c^{hand}}.
    \end{equation}

\end{itemize}

\textbf{Optimization Objective} Previous studies using single-camera setups have indicated that reconstruction inaccuracies primarily stem from incorrect limb assessments, with errors increasing as the distance from the human body to the camera grows due to self-occlusion. To address this, we have integrated the joint distance distribution loss as a spatial prior to refining JRN’s learning process.

\begin{figure}[t]
    \centering
    \includegraphics[width=0.46\textwidth]{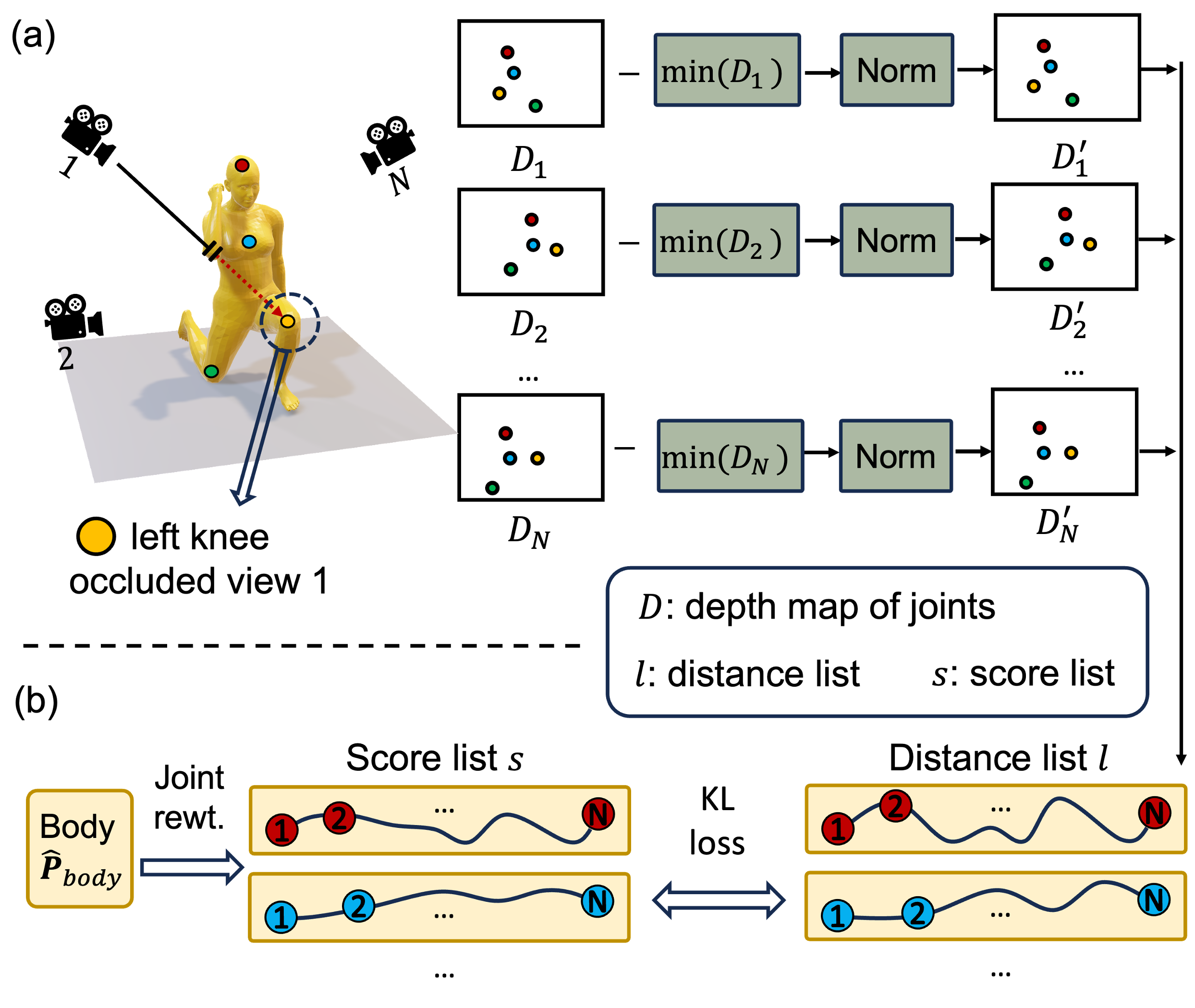}
    \caption{Joint distance distribution loss involves several key steps. First, we subtract the minimal distance value and normalize the depth maps from the ground truth. Then, for the same joint, we align the predicted scores and distance distributions from different camera positions. This alignment enables the model to effectively mitigate the self-occlusion caused by greater distances from the camera position.}
    \label{fig:Rank}
\end{figure}

The optimization process is illustrated in Figure~\ref{fig:Rank}.
Joint distance distribution loss is applied to 21 body joints, but for clarity we draw only 4 joint landmarks in Figure~\ref{fig:Rank}. Initially, 3D joints from view point \(c\) are projected onto a 2D distance map \(D_{c}\in \mathbb{R}^{21\times 3}\) in camera \(c\)'s image coordinate system. 
We subtract the minimal distance value from each distance map and normalize these values to a range from 0 to 1. Through this process, we obtain an approximate representation of the self-occlusion probability distribution.

The distance lists are captured as \( l \in \mathbb{N}^{N \times 21} \). We collect reweighting scores for these joints from \( N \) different cameras (\( s_c^{body} \in \mathbb{R}^{21} \)) into \( s \in \mathbb{R}^{N \times 21} \), aiming for the Joint Reconstruction Network (JRN) to assign higher scores to joints with a lower probability of self-occlusion across views. We then use the Kullback-Leibler divergence to quantify the difference between \( s \) and \( l \). Below is the definition of the joint distance distribution loss function:

\begin{equation}
\label{equ:rank_loss}
    L_{JRN}(s, l) = D_{KL}(s||l).
\end{equation}


\begin{figure}[hpt]
    \centering
    \includegraphics[width=0.5\textwidth]{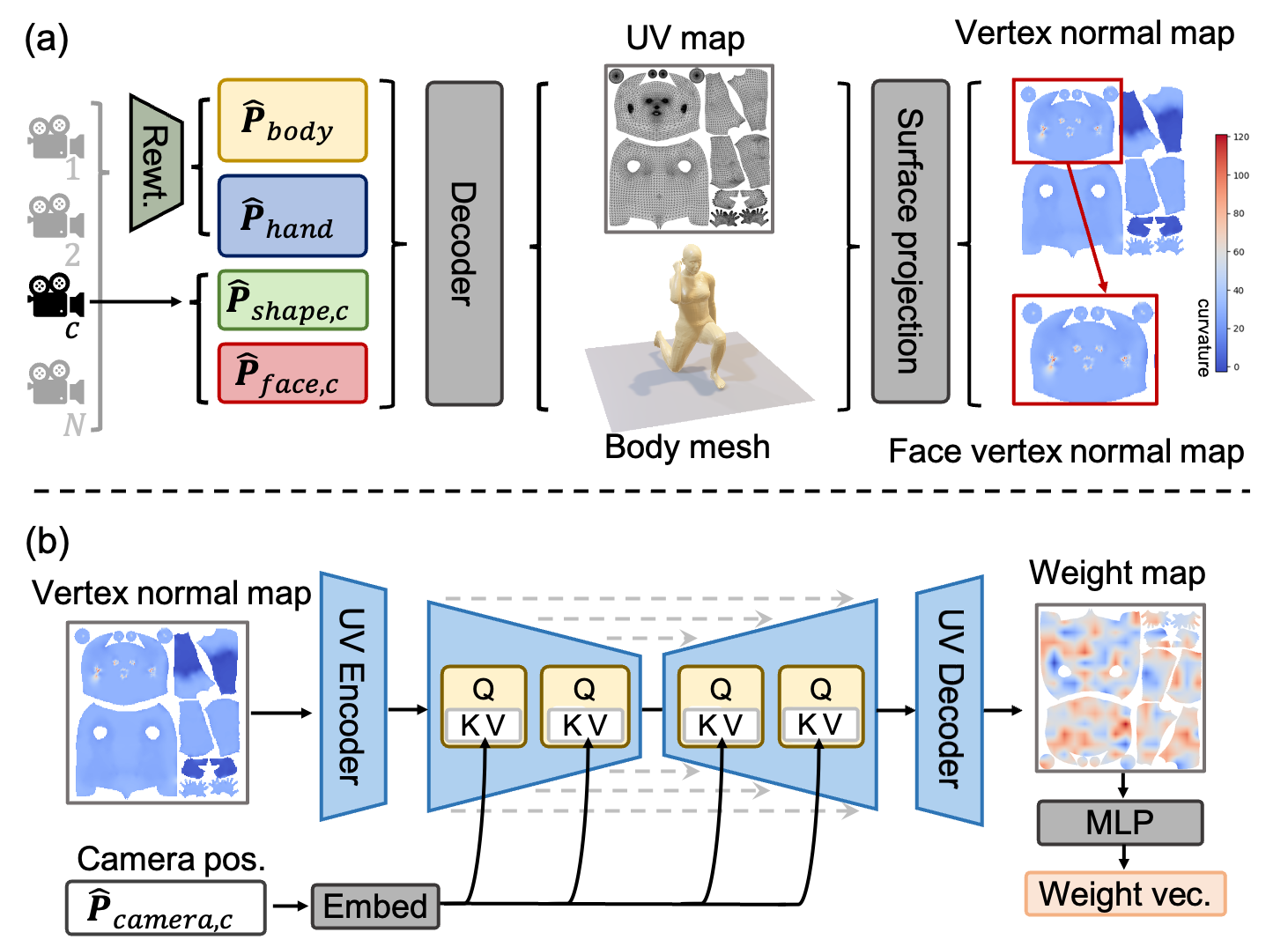}
    \caption{Workflow of the Surface Reweighting Network. (a) The mixed body and hand parameters, together with the shape and expression parameters, are transformed into continuous feature maps through UV projection. (b) Employs the camera position as a condition to facilitate cross-attention operations with the feature map, resulting in the prediction of UV map-level weight maps and PCA-reduced level weight vectors.}
    \label{fig:UV}
\end{figure}


\subsection{Multi-view Surface Reweighting}
\label{sec:sir}
In the last section, we have introduced joint reweighting network, which can mix joint landmarks from multiple cameras. This section introduces the Surface Reweighting Network (SRN), which amalgamates the body surface attached to joint from multiple cameras. 
However, the direct fusion of 10,475 vertices of a 3D body presents a significant challenge for current hardware capabilities and poses difficulties in optimizing the reweighting network. To address this, we propose a novel surface reweighting method that concentrates on the continuous representation of these parameters. This approach facilitates a more refined understanding of the 3D human form. The overview of SRN is illustrated in Figure \ref{fig:UV}.



As shown in Figure \ref{fig:UV} (a), initially, the mixed \(\hat{\mathbf{P}}_{body}\) and \(\hat{\mathbf{P}}_{hand}\) parameters, obtained from the JRN, are inputted into a SMPL-X layer alongside the \(\hat{\mathbf{P}}_{shape, c}\) and \(\hat{\mathbf{P}}_{face, c}\) parameters encoded from view \(c\). This will generate a reconstructed human body mesh, from which vertex normals are projected onto a downsampled UV map. This projection yields the vertex normal map \(VN_{shape, c}\in \mathbb{R}^{U\times V\times 3}\). Similarly, the face vertex normal map \(VN_{face, c}\in \mathbb{R}^{U^{\prime}\times V^{\prime}\times 3}\) can be cropped from \(VN_{shape, c}\).

We employed a conditional U-Net, which is a simplified version akin to the latent diffusion model described by Rombach et al. \cite{rombach2022high}. This U-Net is specifically informed by the predicted camera position \(\hat{\mathbf{P}}_{camera, c}\). The architecture of the model is shown in Figure \ref{fig:UV} (b). Specifically, it contains two Downsample Blocks and two Upsample Blocks with skip connections. For each block, we use \(\hat{\mathbf{P}}_{camera, c}\) as K and V, and features from vertex normal map as Q to compute the cross-attention. This network predicts the weight maps \(W_{shape, c}\in \mathbb{R}^{U\times V\times 3}\) and \(W_{face, c}\in \mathbb{R}^{U\times V\times 3}\). These weight maps are essential for reweighting the body and facial surface features for each camera view, focusing on capturing the continuous variations in body shape and facial expressions. The final \(VN_{shape}\) and \(VN_{face}\) can be obtained by weighted normalization and can be supervised by the real vertex normal map generated by ground truth.

Following this, the predicted weight maps \(W_{shape, c}\) and \(W_{face, c}\) are converted into weight vectors \(w_{shape, c}\in \mathbb{R}^{10}\) and \(w_{face, c}\in \mathbb{R}^{10}\) through MLPs which bring more reasonable \(\hat{\mathbf{P}}_{shape, c}\) and \(\hat{\mathbf{P}}_{face, c}\) reweighting. The final \(\hat{\mathbf{P}}_{shape}\) and \(\hat{\mathbf{P}}_{face}\) can be obtained by weighted normalization as well.

\subsection{Implementation}
We train our model end-to-end by minimizing a several loss functions: $L_{smplx}$, $L_{joint2D}$, $L_{JRN}$, $L_{surface}$. Each component of the loss function serves a specific purpose in the learning process. The \(L_{smplx}\) loss is computed as the L1 distance between the ground truth and predicted SMPL-X parameters, providing explicit supervision for whole body joints, body shape, facial expressions, and camera position. The \(L_{joint2D}\) loss is a regression loss for the projected 2D landmarks of the whole body, ensuring the accuracy of landmark localization in two-dimensional space. The \(L_{JRN}\) loss, as defined in Eq.~\ref{equ:rank_loss}, is used for training JRN to learn the importance of different body joints across various camera views. Finally, the \(L_{surface}\) loss is calculated as the L1 distance between the ground truth and the predicted fused surface feature maps, offering direct supervision for SRN.

All model are trained using single A100 with pytorch. 
Adam optimizer with an initial learning rate of \(3\times 10^{-5}\) for 20 epochs. 
The initialization weight of encoder from SMPLer-X \cite{cai2023smpler}.

\begin{table}[h]
  \centering
    \centering
    \caption{The distribution of camera numbers in the RICH dataset.}
    \resizebox{0.6\linewidth}{!}{
      \begin{tabular}{@{}lccc@{}}
        \toprule
        \multicolumn{1}{c}{Scenes} & Train & Validate & Test \\ \midrule
        BBQ                        & 8     & N/A      & N/A  \\
        Gym                        & 7     & N/A      & 3-4    \\
        LectureHall                & 7     & 7        & 3-4    \\
        ParkingLot 1               & 8     & 8        & N/A  \\
        ParkingLot 2               & 6     & 6        & 3-4    \\
        Pavallion                  & 7     & 7        & N/A  \\ \bottomrule
        \end{tabular}
    }
    \label{tab:rich}
\end{table}

\begin{table*}[h]
\caption{Comparison between the proposed method and existing calibration-free single view and multiview methods on Human3.6M dataset. 
Since multiview reconstruction is not supported by single view methods, we report the mean of results from each view. 
}
\centering
\scalebox{1}{
\begin{tabular}{lcccccc}
\hline
\multirow{2}{*}{Method}                    & \multirow{2}{*}{Camera(s)} & \multirow{2}{*}{Model type} & \multicolumn{2}{c}{Multi-view reconstruction} & \multicolumn{2}{c}{Single-view reconstruction} \\ \cline{4-7} 
                                           &                            &                             & PA-MPJPE              & PA-MPVPE              & PA-MPJPE               & PA-MPVPE              \\ \hline
SMPLify (\citeyear{bogo2016keep})          & Single                     & SMPL                        & N/A                   & N/A                   & 82.3                   & N/A                   \\
HMR (\citeyear{kanazawa2018end})           & Single                     & SMPL                        & 57.8                  & 67.7                  & 56.8                   & 65.5                  \\
GraphCMR (\citeyear{kolotouros2019convolutional}) & Single              & SMPL                        & 50.9                  & 59.1                  & 50.1                   & 56.9                  \\
SPIN (\citeyear{kolotouros2019learning})   & Single                     & SMPL                        & 44.5                  & 51.5                  & 41.1                   & 49.3                  \\
DecoMR (\citeyear{zeng20203d})             & Single                     & SMPL                        & 42.0                  & 50.5                  & 39.3                   & 47.6                  \\
Pose2Mesh (\citeyear{choi2020pose2mesh})   & Single                     & SMPL                        & N/A                   & N/A                   & 47.0                   & N/A                   \\
I2lMeshnet (\citeyear{moon2020i2l})        & Single                     & SMPL                        & N/A                   & N/A                   & 41.1                   & N/A                   \\
PyMAF (\citeyear{zhang2021pymaf})          & Single                     & SMPL                        & N/A                   & N/A                   & 40.5                   & N/A                   \\
SMPLer-X-base (\citeyear{cai2023smpler})   & Single                     & SMPL-X                      & 38.3                  & 41.3                  & 45.1                   & 47.8                  \\
SMPLer-X-large (\citeyear{cai2023smpler})  & Single                     & SMPL-X                      & 33.4                  & 37.1                  & \textbf{38.9}          & \textbf{42.8}         \\ \hline
Liang (\citeyear{liang2019shape})          & Multi                      & SMPL                        & 48.5                  & 57.5                  & 59.1                   & 69.2                  \\
ProHMR (\citeyear{kolotouros2021probabilistic})   & Multi               & SMPL                        & 34.5                  & N/A                   & 41.2                   & N/A                   \\
Yu (\citeyear{yu2022multiview})            & Multi                      & SMPL                        & 33.0                  & \ul{ 34.4}            & 41.6                   & 46.4                  \\
PaFF (\citeyear{jia2023delving})           & Multi                      & SMPL                        & \ul{ 28.2}            & N/A                   & N/A                    & N/A                   \\
MUC-base                                  & Multi                      & SMPL-X                      & 31.9                  & \textbf{33.4}         & 44.3                   & \ul{ 45.8}            \\
MUC-base (with translation)               & Multi                      & SMPL                        & \textbf{27.1}         & N/A                   & \ul{ 39.5}              & N/A                   \\ \hline
\end{tabular}
}
\label{tab:compare}
\end{table*}
\section{Experiments}
\label{sec:exp}

In our experiments, we demonstrate the effectiveness and robustness of our proposed MUC method for calibration-free multi-view fusion. 
\begin{itemize}
    \item Initially, we benchmark our method against state-of-the-art 3D human body reconstruction techniques, providing a detailed comparison on accuracy and robustness.
    \item Subsequently, we delve into the core of our contributions, evaluating the performance with an increasing number of cameras. This involves a systematic analysis to illustrate the scalability and impact of multi-view geometry on reconstruction accuracy.
    \item Lastly, we conduct an ablation study to reveal the significance of our proposed JRN and SRN based on joint distribution loss, isolating their effects and demonstrating their contributions to the overall performance.
\end{itemize}

\subsection{Datasets and Metrics}
Two datasets are used to train and evaluate our method. 
\begin{itemize}
    \item \textbf{Human3.6M}~\cite{ionescu2013human3} is a large-scale multi-view dataset with ground-truth 3D human pose annotation. 
    We follow the standard training/testing split: using subjects S1, S5, S6, S7 and S8 for training, and subjects S9 and S11 for testing. 
    The SMPL/SMPL-X pseudo-GTs are obtained from NeuralAnnot \cite{moon2022neuralannot}. 
    This dataset is widely used in multiple human body reconstruction tasks and helps us to compare with other state-of-the-art methods.
    \item \textbf{RICH}~\cite{huang2022capturing} is an in-the-wild and indoor multi-view dataset annotated with ground-truth 2D keypoints and 3D body mesh. 
    We adopt the intrinsic training/testing split in the dataset. 
    The SMPL-X pseudo-GTs are obtained from SMPLify-X. 
    This dataset contains human body joints, shape, hand joints and facial expression annotations, which can be used to prove the effectiveness of our multi-view fusion method. The distribution of camera numbers is shown in Table \ref{tab:rich}. For each sample in the training and validation sets, we randomly split it into two samples, each with a camera count equal to 4.
\end{itemize}

Both of these datasets are based on video capture, where the positions of individuals relative to the cameras are constantly changing, hence it can be considered suitable for assessing the generalization ability of our method to camera positions.

For 3D whole-body mesh reconstruction, we utilize the mean per-vertex position error (MPVPE) and mean per-joint position error (MPJPE) as our primary metrics. 
In addition, we apply procrustes analysis (PA) to the recovered mesh, and report PA-MPVPE and PA-MPJPE after rigid alignment. 
PA-MPJPE primarily evaluates the pose estimation accuracy of the human body model, whereas PA-MPVPE evaluates the accuracy of the reconstructed 3D model. PA-MPJPE and PA-MPVPE are reported in millimeter (mm).

\begin{figure*}
  \centering
  \includegraphics[width=0.9\textwidth]{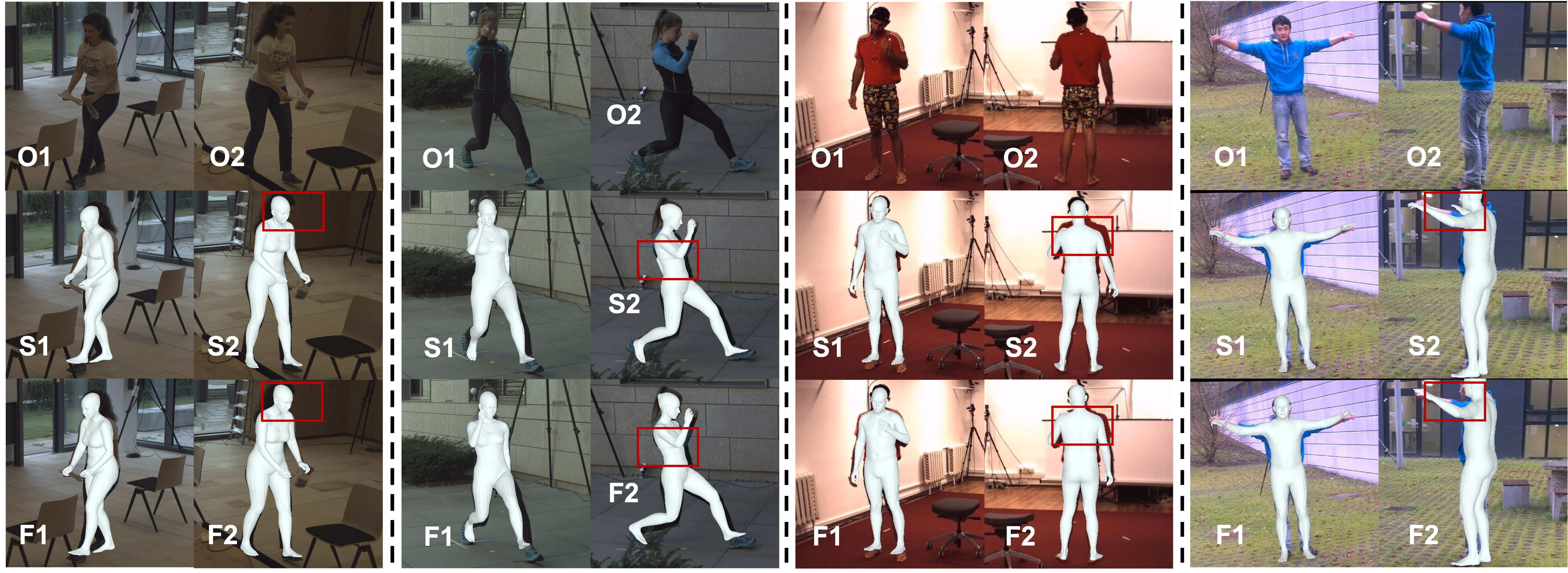}
  \caption{We conducted a qualitative comparison with SMPLer-X across three datasets. The first two groups are from the RICH dataset, the third group is from the Human3.6M dataset, and the last group is from an additional validation dataset, the MARCOnI dataset \cite{elhayek2015efficient}, which serves as a more challenging test scenario (images were recorded using a handheld smartphone).
  ``O'' stands for the original image. 
  ``S'' stands for single-view reconstruction result by SMPLer-X. 
  ``F'' stands for the fusion result of our method. Zoom in for better view.}
  \label{fig:mono-vs-mutiple}
\end{figure*}

\subsection{Comparative Analysis with SOTA Methods}

In this subsection, we present a comprehensive comparison of our proposed calibration-free multi-view fusion model with current SOTA methods in the domain of 3D human body reconstruction. 
We employ both multi-view and single-view reconstruction methods on the Human3.6M dataset to gauge the performance of our method against an array of established techniques.
The results are presented in Table~\ref{tab:compare}.

Our comparison includes 3D human body reconstruction methods from 2016 to 2023.
Most of them are in the category of single view, while a few are in the multi-view category. 
Meanwhile, many methods adopt the SMPL model, which can only represent human body pose and body shape.
A more sophisticated SMPL-X, as adopted by our method, can reconstruct human body pose, hand pose, body shape and facial expression  jointly. 
Specifically, we can treat SMPLer-X \cite{cai2023smpler} as SOTA method for single-view fusion method and PaFF \cite{jia2023delving} as SOTA method for multi-view fusion method. To eliminate differences between using different templates, we not only evaluated using the SMPL-X template but also utilized the conversion tool provided by SMPL-X~\cite{pavlakos2019expressive} to transform our prediction results into the SMPL template for evaluation.

From Table~\ref{tab:compare}, we can observe the following.
\begin{itemize}
    \item Our Method's Edge: Our approach outperforms other methods in multi-view reconstruction, achieving a PA-MPJPE of 27.1mm, which surpasses the previous best multi-view method by 1.1mm. It also performs better than models with larger parameter counts, such as SMPLer-X-large. Additionally, it maintains a leading position in both PA-MPJPE and PA-MPVPE metrics in single-view scenarios.
    \item Model Bias: Under different model types (SMPL and SMPL-X), our approach demonstrates significant performance improvements in multi-view reconstruction compared to single view. Moreover, it achieves a notable enhancement in joints accuracy compared to previous methods. Since our model utilizes the SMPL-X model for both training and prediction, and the vertex count of the SMPL-X model is roughly twice that of the SMPL model, forcibly converting results to the SMPL model introduces some errors in the overall vertex alignment. Therefore, we only present results in terms of PA-MPJPE.
\end{itemize}

From Table~\ref{tab:rich_compara}, we can observe that our method outperforms SMPLer-X on the RICH dataset, which features more complex scenes and diverse camera settings. Our method's visualization results are depicted in Figure~\ref{fig:mono-vs-mutiple}. Compared to single-view models, our approach better compensates for the information loss caused by different viewpoints, resulting in superior reconstruction outcomes.

\begin{table}[h]
  \centering
    \caption{Comparison with SMPLer-X on RICH dataset. Since SMPLer-X is a single view method, we report the mean of results from each view.}
    \resizebox{1\linewidth}{!}{
      \begin{tabular}{lcc}
    \hline
    \multicolumn{1}{c}{\multirow{2}{*}{Method}} & \multicolumn{2}{c}{Multi-view}                              \\
    \multicolumn{1}{c}{}                        & \multicolumn{1}{l}{PA-MPJPE} & \multicolumn{1}{l}{PA-MPVPE} \\ \hline
    SMPLer-X-base (\citeyear{cai2023smpler})    & 42.1                         & 37.1                         \\
    MUC-base                                   & \textbf{37.5}                & \textbf{33.5}                \\ \hline
    \end{tabular}
    }
    \label{tab:rich_compara}
\end{table}

\subsection{Performance across Multi-view Configurations}

\begin{figure}[t]
  \centering
  \includegraphics[width=0.9\linewidth]{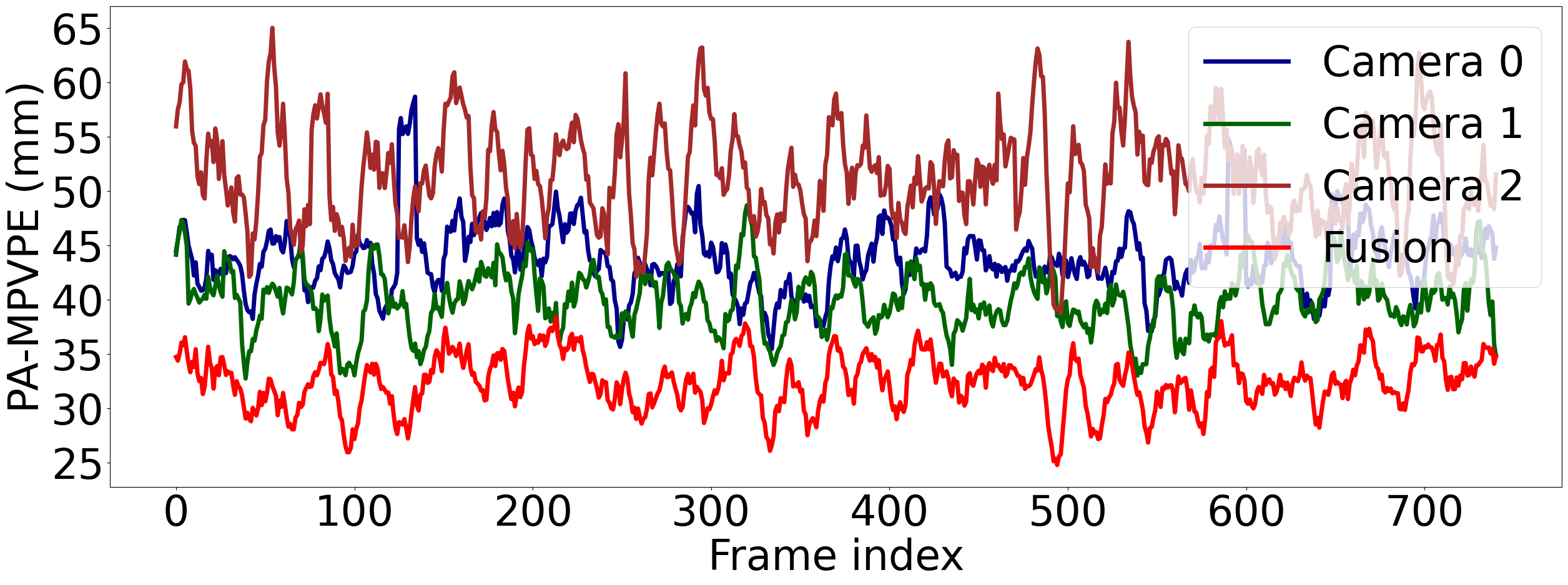}
  \caption{Temporal comparison of PA-MPVPE across different camera setups over sequential frames. 
  Mono-view reconstructions from cameras 0, 1, and 2 are depicted in varying colors, while the multi-view reconstruction is represented in red. Zoom in for better view.}
  \label{fig:across_time}
\end{figure}

To evaluate the effectiveness of our calibration-free multi-view fusion method, we first present a frame-based analysis for a video of three cameras. 
As illustrated in Figure~\ref{fig:across_time}, our multi-view method effectively compensates for the target's movement in the wild, maintaining a stable PA-MPVPE fluctuation. 

Our method is scalable to an arbitrary number of cameras out of the box, so we conduct quantitative validation to assess the impact of adding more cameras for each case. The goal is to determine how the number of cameras influences the reconstruction precision of different human body parts. 

The results, shown in Tables \ref{tab:ncam}, underscore our method's effectiveness in utilizing additional camera angles. For body reconstruction, there is a marked decrease in PA-MPJPE when increasing the camera count from one to four. Similar enhancements are observed in the hand and face reconstructions. Notably, the transition from one to two cameras yields the most significant improvement, highlighting the value of multi-view geometry. While the incremental gains diminish with each additional camera, the consistent improvements across all metrics reinforce our approach's effectiveness in leveraging multiple uncalibrated views for 3D human body reconstruction in various settings. The same conclusions are also demonstrated on the Human3.6M dataset, with results presented in Table \ref{tab:ncam_h36m}.

\begin{table}[]
\centering
\caption{Different camera number on RICH dataset. We select cases contain 4 cameras from the test set to evaluate our method with different camera numbers.}
\resizebox{\linewidth}{!}{
\begin{tabular}{c|ccccc}
\hline
\multirow{2}{*}{\# Cam} & \multicolumn{2}{c}{Body}      & \multicolumn{2}{c}{Hand}    & Face         \\
                        & PA-MPJPE      & PA-MPVPE      & PA-MPJPE     & PA-MPVPE     & PA-MPVPE     \\ \hline
1                       & 52.8          & 47.7          & 8.4          & 8.2          & 4.1          \\
2                       & 43.7          & 38.9          & 7.4          & 7.2          & 3.5          \\
3                       & \ul{ 40.4}    & \ul{ 35.8}    & \ul{ 7.0}    & \ul{ 6.8}    & \ul{ 3.3}    \\
4                       & \textbf{38.6} & \textbf{34.5} & \textbf{6.8} & \textbf{6.7} & \textbf{3.2} \\ \hline
\end{tabular}
}
\label{tab:ncam}
\end{table}

\begin{table}[]
  \centering
  \caption{Different camera number on Human3.6M dataset. }
  \resizebox{\linewidth}{!}{
  \begin{tabular}{c|ccccc}
  \hline
  \multirow{2}{*}{\# Cam} & \multicolumn{2}{c}{Body}      & \multicolumn{2}{c}{Hand}    & Face         \\
                          & PA-MPJPE      & PA-MPVPE      & PA-MPJPE     & PA-MPVPE     & PA-MPVPE     \\ \hline
  1                       & 47.0          & 49.2          & 1.1          & 1.3          & 5.0          \\
  2                       & 37.1          & 38.8          & \ul{ 0.9}    & \ul{ 1.0}    & 4.2          \\
  3                       & \ul{ 34.5}    & \ul{ 36.4}    & \ul{ 0.9}    & \textbf{0.9} & \ul{ 4.1}    \\
  4                       & \textbf{31.9} & \textbf{33.4} & \textbf{0.8} & \textbf{0.9} & \textbf{3.9} \\ \hline
  \end{tabular}
  }
  \label{tab:ncam_h36m}
  \end{table}

\subsection{Impact of Joint and Surface Reweighting}
\begin{figure}[t]
  \centering
  \includegraphics[width=1\linewidth]{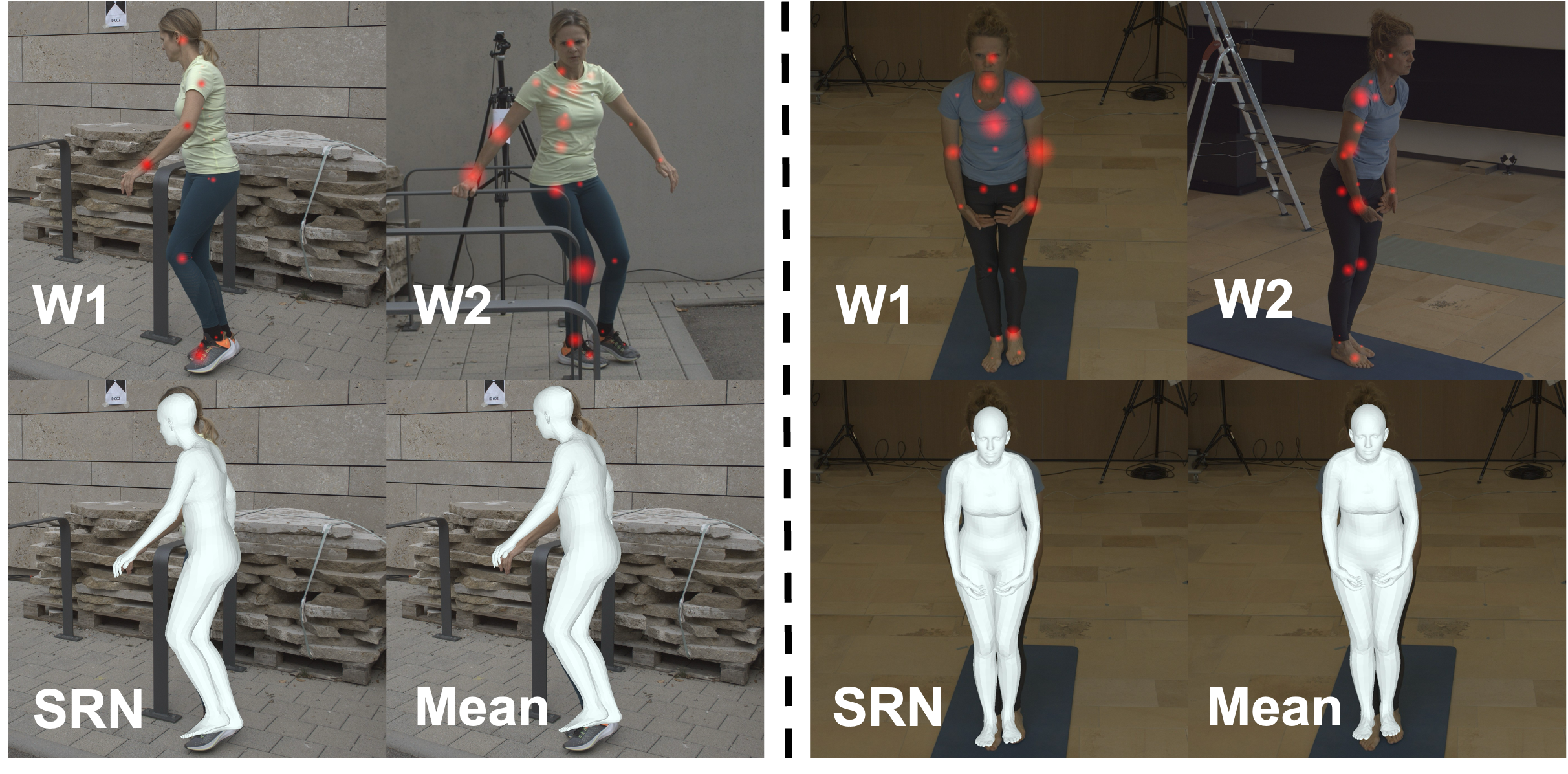}
  \caption{Visualization of the predicted scores by JRN across different views, accompanied by an ablation study on SRN. Larger red circles represent higher weights. For clarity, only 21 body joints are selected for display. 
  ``W'' stands for the visualization of joint weight.}
  \label{fig:jrn}
\end{figure}

Our qualitative evaluation focuses on the influence of the JRN and SRN on the 3D reconstruction of human body poses and shape. Figure~\ref{fig:jrn} visually demonstrates the effect of reweighting across different camera views.

In Figure~\ref{fig:jrn}, the size of the red circles superimposed on the model's joints indicates the importance level that JRN assigns to each joint. Larger circles indicate a higher score, suggesting that these joints exert a more significant influence during the multi-view joint reweighting process. It is evident that with the aid of the joint distance distribution loss, the JRN assigns lower weights to joints with a higher probability of self-occlusion. This strategy effectively harnesses the more accurate parts of predictions from each single viewpoint for fusion. Similarly, predictions using the SRN feature more precise estimates of body shape.

\begin{table}[h]
  \centering
    \centering
    \caption{Ablation on JRN and SRN for whole body reconstruction. Best results are marked in bold.}
    \resizebox{0.98\linewidth}{!}{
    \begin{tabular}{cc|cccc}
    \toprule
      &    & \multicolumn{2}{c}{Human3.6M} & \multicolumn{2}{c}{RICH} \\
    JRN & SRN & PA-MPJPE      & PA-MPVPE      & PA-MPJPE    & PA-MPVPE   \\\midrule
      &    & 33.8         & 34.8         & 42.1       & 37.1      \\                
    \Checkmark &    & \ul{32.5}      & \ul{34.3}       & \ul{39.4}    & \ul{34.2}      \\       
    \Checkmark & \Checkmark & \textbf{31.9}          & \textbf{33.4}          & \textbf{37.5}       & \textbf{33.5}    \\
    \bottomrule

    \end{tabular}
    }
    \label{tab:body}
  \end{table}

  \begin{table}[t]
    \centering
    \label{tab:hand}
    \caption{Ablation on JRN and SRN for hand and face on RICH dataset. Best results are marked in bold.}
    \resizebox{0.78\linewidth}{!}{
    \begin{tabular}{cc|ccc}
    \toprule
      &    & \multicolumn{2}{c}{Hand} & Face     \\
    JRN & SRN & PA-MPJPE    & PA-MPVPE   & PA-MPVPE \\ \midrule
              &            & 7.2       & 7.0      & 5.0     \\      
    \Checkmark &            & \underline{7.0}       & \underline{6.8}      & \underline{3.5}     \\      
    \Checkmark & \Checkmark & \textbf{6.9}       & \textbf{6.7}      & \textbf{3.3}     \\ \bottomrule
    \end{tabular}
    }

    \label{tab:handface}
\end{table}

We further perform two quantitative ablation studies experiments to objectively measure the effectiveness of the JRN and SRN: one focusing on whole body reconstruction and the other on detailed hand and face reconstruction.

\textbf{Whole Body Reconstruction} We utilized the Human3.6M and RICH datasets to evaluate the performance improvements. The results are compiled in Table~\ref{tab:body}. The inclusion of JRN alone, and in combination with SRN, shows a consistent decrease in PA-MPJPE and PA-MPVPE across both datasets, highlighting the benefits of the reweighting mechanisms.

\textbf{Hand and Face Reconstruction} The results on RICH are presented in Table~\ref{tab:handface}. The quantitative analysis clearly demonstrates that the combined use of JRN and SRN yields the best performance, as indicated by the bold values for PA-MPJPE and PA-MPVPE, underscoring their importance in enhancing the precision of 3D human pose reconstruction.

\section{Conclusion and Discussion}
\label{sec:conclusion}
In this paper, we present a new method for creating 3D meshes from images taken by uncalibrated cameras, removing the need for complex setup. Our approach scales easily to any number of cameras and uses distance distribution to estimate self-occlusion, enhanced by our SRN for smooth surface fusion.
The technique determines the reliability of image segments to improve 3D reconstruction and assesses the positioning confidence of body parts from different views. Currently limited to static images, future updates could extend to video for more precise pose estimation.
This method is not only more practical for everyday use but also surpasses current leading techniques in accuracy.


%
%

\bibliography{aaai25}


\end{document}